\documentclass{article}
\usepackage{ijcai26}

\usepackage{times}
\usepackage{soul}
\usepackage{url}
\usepackage[hidelinks]{hyperref}
\usepackage[utf8]{inputenc}
\usepackage[small]{caption}
\usepackage{graphicx}
\usepackage{amsmath}
\usepackage{amsthm}
\usepackage{booktabs}
\usepackage{algorithm}
\usepackage{algorithmic}

\usepackage{array}  
\usepackage{pifont}
\usepackage{amssymb}   
\usepackage{dblfloatfix}  

\usepackage{fancyhdr}
\makeatletter
\def\blfootnote{\gdef\@thefnmark{}\@footnotetext}
\makeatother

\title{LFS: Learnable Frame Selector for Event-Aware and Temporally Diverse Video Captioning}
\author{
	Lianying Chao\textsuperscript{\textnormal{1}}, Linfeng Yin\textsuperscript{\textnormal{1}}, Peiyu Ren, Fanyi Jiang, Qiaoyu Ren, Dingcheng Shan, Jingcheng Pang, Sijie Wu, Xubin Li\textsuperscript{\textnormal{*}}, Kai Zhang\textsuperscript{\textnormal{*}}
	\vspace{1mm}, Xin Chen
	\affiliations
	GTS, AI Data Department, Huawei Technologies Co., Ltd. 
	\vspace{1mm}
	\emails
	{\{lixubin,zhangkai304\}}@huawei.com
	\vspace{1mm}
	\affiliations
}

\begin{document}
	\maketitle
	\thispagestyle{fancy} 

	\begin{abstract}
		Video captioning models convert frames into visual tokens and generate descriptions with large language models (LLMs). Since encoding all frames is prohibitively expensive, uniform sampling is the default choice, but it enforces equal temporal coverage while ignoring the uneven events distribution. This motivates a Learnable Frame Selector (LFS) that selects temporally diverse and event-relevant frames. LFS explicitly models temporal importance to balance temporal diversity and event relevance, and employs a stratified strategy to ensure temporal coverage while avoiding clustering. Crucially, LFS leverages caption feedback from frozen video-LLMs to learn frame selection that directly optimizes downstream caption quality. Additionally, we identify the gap between existing benchmark and human's cognition. Thus, we introduce ICH-CC built from carefully designed questions by annotators that reflect human-consistent understanding of video. Experiments indicate that LFS consistently improves detailed video captioning across two representative community benchmarks and ICH-CC, achieving up to 2.0\% gains on VDC and over 4\% gains on ICH-CC. Moreover, we observe that enhanced captions with LFS leads to improved performance on video question answering. Overall, LFS provides an effective and easy-to-integrate solution for detailed video captioning.
	\end{abstract}

	\section{Introduction}
	
	Recent advances in multimodal large language models (MLLMs) have greatly improved visual-to-text generation by encoding visual inputs as tokens processed by LLMs \cite{10386743}. Compared with image-related tasks, video understanding tasks requires jointly modeling spatial content and temporal dynamics \cite{zheng2023judging,10657734,chen2024longvila}, and are commonly categorized into video question answering (QA) \cite{TIAN2025104215}, short captioning \cite{10814098}, and detailed captioning \cite{chai2025auroracap}. Among them, the latter targets fine-grained and temporally coherent descriptions, providing richer information than the former two and better supporting downstream tasks such as video retrieval and multimodal QA. Accordingly, this work focuses on detailed video captioning, and Fig.~1 demonstrates consistent improvements over baselines across nine benchmarks spanning both detailed captioning and QA tasks.
	
	\begin{figure}[t]
		\centering
		\includegraphics[width=0.45\textwidth]{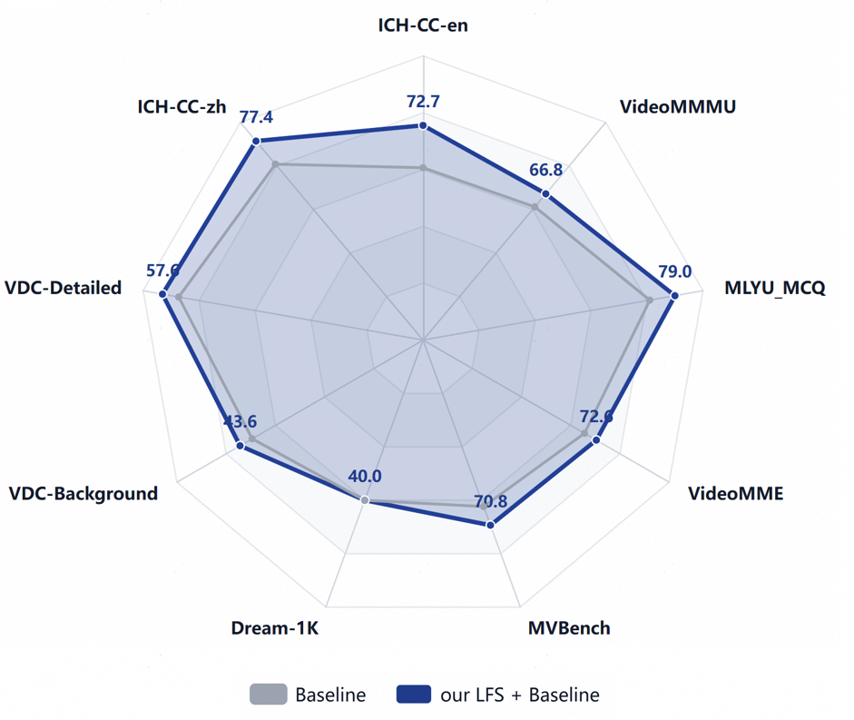}
		\caption{Performance comparison between baselines and the LFS-enhanced counterparts across nine benchmarks. Baselines include AuroraCap-7B, Tarsier2-7B, Qwen2.5-VL-7B, and Qwen3-VL-8B.}
		\label{fig:sample}
	\end{figure}
	
	Due to computational constraints, most video-LLMs adopt uniform sampling to select a fixed number of frames \cite{Maaz2023VideoChatGPT,li2024videochat,lin2023videoLLAVA}. While this strategy ensures temporal coverage, it ignores the uneven distribution of informative events and often underrepresents short but critical actions. In contrast, query-driven frame selection methods for video QA retrieve frames relevant to a given question \cite{tang2025AKS,Zhu2025FOCUSEK,zhang2025qframe}, but are unsuitable for query-agnostic tasks such as detailed video captioning.
	
	An effective frame selection strategy for detailed captioning should satisfy two requirements: (1) event awareness, capturing salient actions and scene changes, and (2) temporal diversity, avoiding redundant frames concentrated in short intervals. However, existing approaches typically fail to meet both: Top-$K$ selection leads to temporal clustering, while reinforcement learning-based methods introduce high variance and are difficult to integrate with frozen video-LLMs.
	
	To address these limitations, we propose a Learnable Frame Selector (LFS) that selects temporally diverse and event-relevant frames. LFS integrates an event-aware temporal scoring network with a stratified Top-$K$ selection mechanism to balance event relevance and temporal diversity. Furthermore, LFS leverages caption feedback from frozen video-LLMs as supervision, directly optimizing frame selection for caption quality rather than proxy objectives, and can be seamlessly integrated into existing pipelines without modifying language model parameters.
	
	In the evaluation of detailed video captioning, representative benchmarks such as VDC and Dream-1K report relatively low performance. Prior studies further show that even experienced human annotators achieve accuracies below 50\% \cite{Dream-1K,chai2025auroracap}, highlighting a substantial gap between benchmark evaluation and human understanding. To address this gap, we introduce ICH-CC, a benchmark for detailed video captioning on intangible cultural heritage cuisine videos, constructed with fully human-authored captions and carefully designed question--answer pairs.
	
	Experiments demonstrate that LFS significantly improves performance on ICH-CC, boosting Qwen3-VL by 3.18\% and 4.47\% on the Chinese and English subsets, respectively. On community benchmarks such as VDC, LFS consistently improves multiple video-LLM backbones under saturated evaluation settings. Moreover, gains in detailed captioning achieved by LFS reliably transfer to downstream video QA.
	
	In summary, frame selection is a critical bottleneck for detailed video captioning. By explicitly modeling event relevance and temporal diversity, LFS provides an effective and easily deployable solution that improves captioning quality. Our main contributions are:
	\begin{itemize}
		\item We propose LFS, a learnable frame selector that balances event awareness and temporal diversity via stratified Top-$K$ selection and caption-guided supervision.
		\item We introduce ICH-CC, a benchmark aligned with human cognition for evaluating detailed video captioning in Chinese and English.
		\item We conduct extensive experiments and show that LFS consistently improves detailed video captioning and zero-shot video QA with multiple video-LLMs and benchmarks.
	\end{itemize}

	\begin{figure*}[t]
		\centering
		\includegraphics[width=\textwidth]{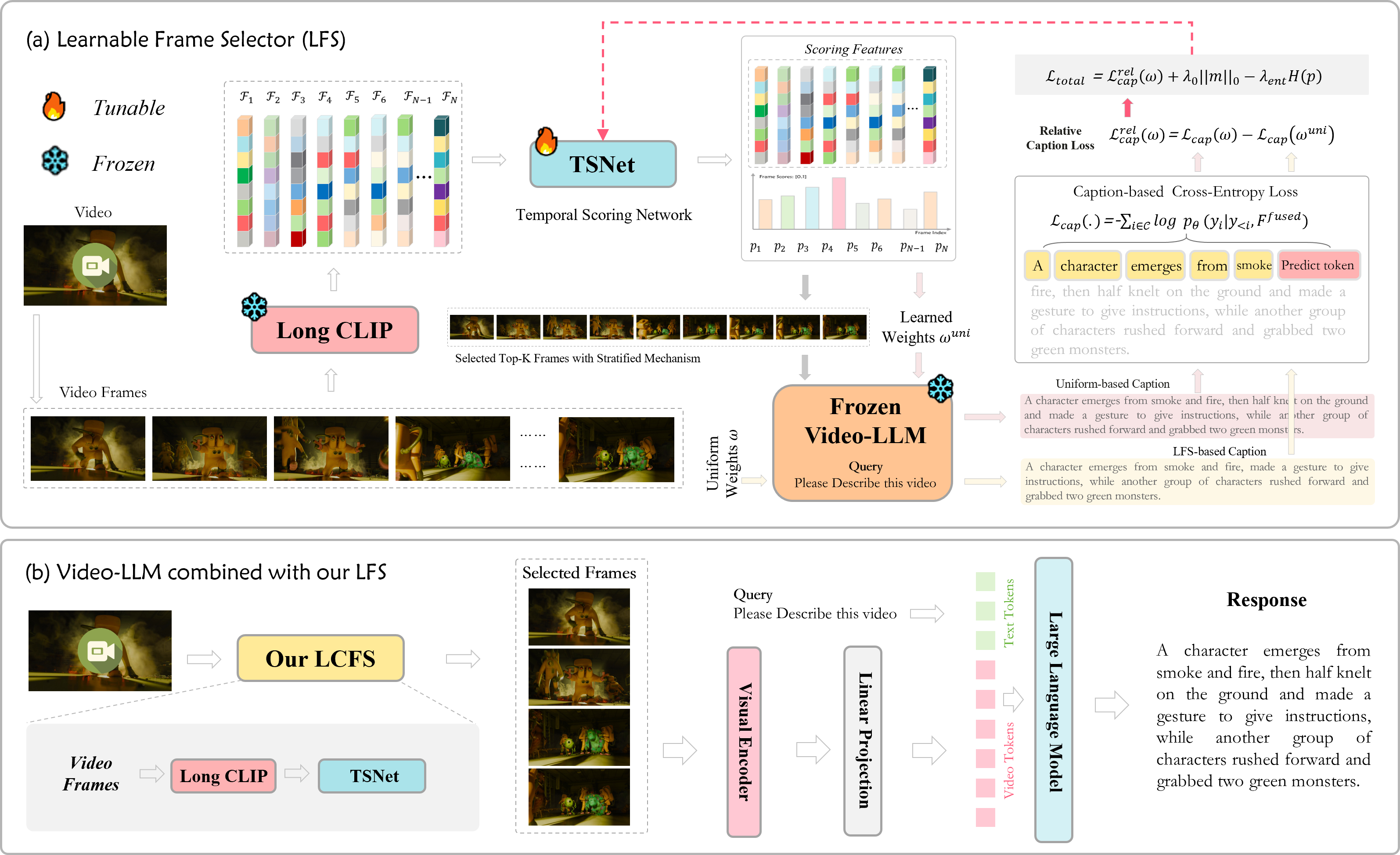}
		\caption{The overall scheme of the proposed learnable frame selector (LFS). (a) and (b) are the training and inference process of LFS, respectively.}
		\label{fig:sample}
	\end{figure*}

	\section{Related Work}
	
	\subsection{Video-LLMs for Question Answering}
	
	Early Video-LLMs, such as Video-Chat \cite{li2024videochat} and Video-LLaMA \cite{damonlpsg2023videollama}, concatenate frame features with text prompts and fine-tune LLMs on instruction-tuning datasets, enabling open-ended dialogue and basic temporal reasoning but suffering from long-context inefficiency. Subsequent works improve scalability through spatio-temporal pooling, Q-former adapters \cite{cite1,10350541,kim2024qformer}, and token dropping, while specialized pre-training objectives enhance motion and event understanding. Despite these advances, most Video-LLMs remain optimized for query-driven reasoning, leaving frame selection for query-agnostic tasks relatively underexplored.
	
	\subsection{Video-LLMs for Detailed Captioning}
	
	Recent work extends video captioning from brief summaries to detailed descriptions \cite{Kim_2025_ICCV,Wu_2025_CVPR,AAAI}. Models such as Video-LLaVA \cite{lin2023videoLLAVA}, VideoChat2 \cite{li2024videochat}, and PLLaVA \cite{xu2024pllava} are fine-tuned on large-scale video--text datasets and evaluated using standard captioning metrics, but often generalize poorly to fine-grained and temporally structured descriptions. To mitigate this issue, benchmarks such as Dream-1K \cite{Dream-1K} and VDC \cite{chai2025auroracap} have been introduced, along with methods like AuroraCap that reduce visual tokens via token merging. Nevertheless, most approaches still rely on uniform frame sampling, which frequently misses short yet informative events in long videos.
	
	\subsection{Frame Selection before Video-LLMs}
	
	To reduce computational cost, prior frame selection methods aim to retain task-relevant frames. Rule-based approaches use uniform subsampling or shot boundary detection, while model-based methods typically select Top-$K$ frames based on query relevance using CLIP-like models \cite{guo2025logic,Hu2025,tang2025tspo}. Token-level selectors further prune redundant visual tokens within frames. While effective for query-driven tasks such as video QA, these methods often overlook temporal diversity and are not designed for query-agnostic settings \cite{li2025maxinfo,11093988}. In contrast, detailed video captioning requires selecting event-aware and temporally-diverse frames that capture distinct events across the entire video, a requirement largely unaddressed by existing frame selection approaches.

	\section{Method}
	
	\subsection{Overall Framework}
	
	We propose Learnable Frame Selector (LFS), a lightweight and plug-and-play module that selects a small set of informative and temporally diverse frames for detailed video captioning, as shown in Fig.2.
	
	Given frozen frame embeddings extracted by a vision encoder Long-CLIP \cite{zhang2024longclip}, LFS predicts a continuous temporal importance field over all frames using a lightweight temporal scoring network (TSNet). Unlike uniform sampling or query-dependent retrieval, LFS models frame importance in a query-agnostic manner. At inference, a stratified Top-$K$ strategy enforces temporal coverage and avoids frame clustering. During training, LFS is optimized with caption-guided supervision from a frozen video-LLM as captioner, directly aligning frame selection with caption quality. 
	
	\subsection{Temporal Importance Modeling}
	
	Let $X=\{x_t\}_{t=1}^{N}$, $x_t \in \mathbb{R}^{d}$ denote frame embeddings from the frozen Long-CLIP. LFS predicts an importance logit $s(t)$ for each frame using TSNet shown in Fig.3.
	
	TSNet first applies a temporal convolution to capture local transitions:
	\begin{equation}
		H_1 = \mathrm{GELU}\!\left(\mathrm{Conv}_1(X)\right), \qquad
		H_1 \in \mathbb{R}^{\text{hid}\times N}.
	\end{equation}
	A global summary $g=\mathrm{Mean}(H_1)$ is used for gated modulation:
	\[
	H_1 \leftarrow H_1 \odot \left(1 + \alpha \tanh(\mathrm{MLP}(g))\right)
	\]
	which adaptively rescales temporal features while preserving near-identity behavior at initialization.
	
	A second temporal convolution aggregates local responses:
	\begin{equation}
		H_2 = \mathrm{GELU}\!\left(\mathrm{Conv}_2(H_1)\right)
	\end{equation}
	followed by a pointwise projection:
	\begin{equation}
		s(t) = \mathrm{Conv}_{1\times1}(H_2)_t
	\end{equation}
	Finally, logits are normalized within each video:
	\begin{equation}
		\hat{s}(t)=\frac{s(t)-\mu_s}{\sqrt{\sigma_s^2+\epsilon}}
	\end{equation}
	
	\begin{figure}[htbp]
		\centering
		\includegraphics[width=0.45\textwidth]{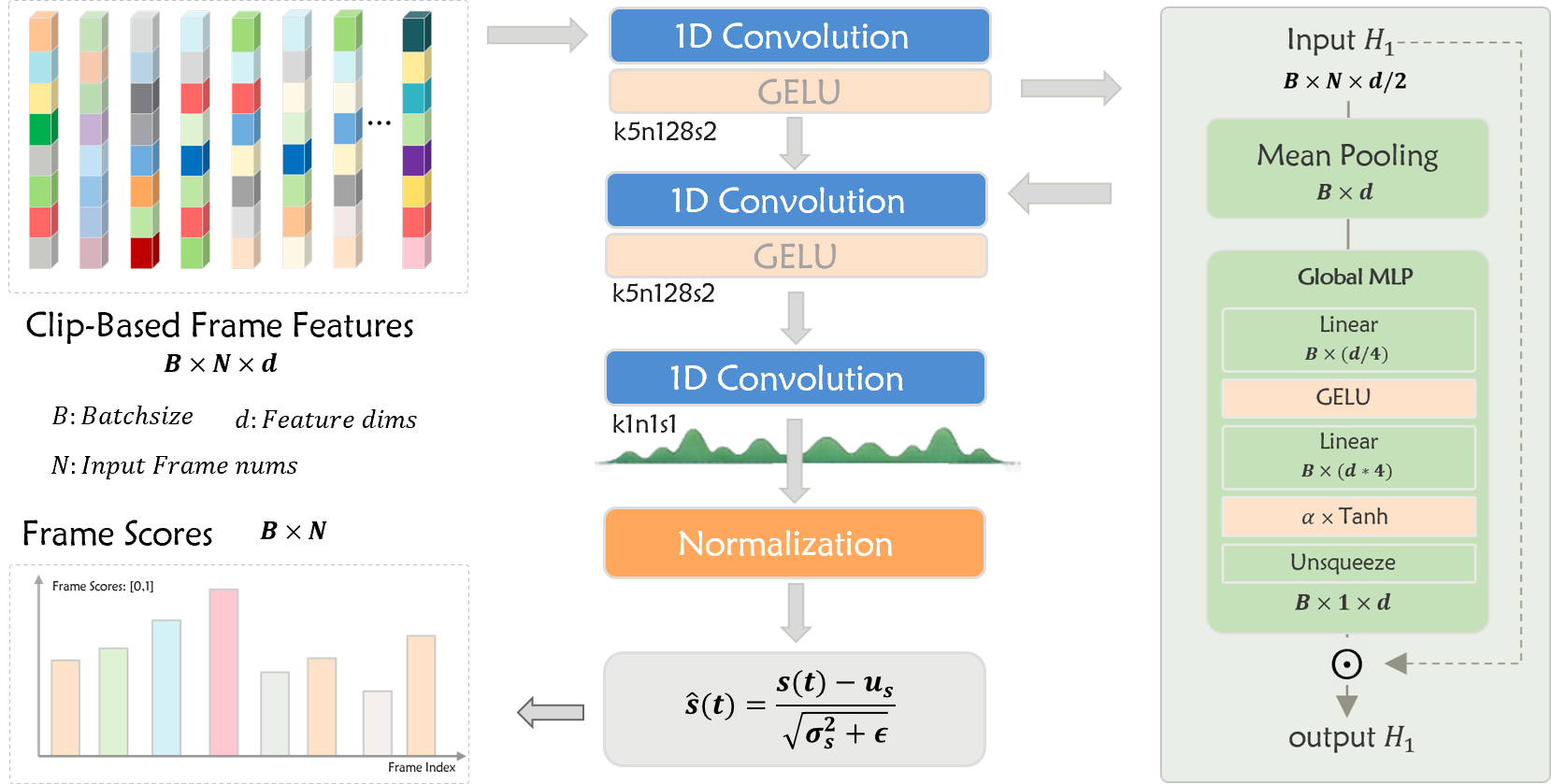}
		\caption{The architecture of temporal scoring network (TSNet).}
		\label{fig:sample}
	\end{figure}

	\subsection{Stratified Top-$K$ Frame Selection}
	
	The normalized logits $\hat{s}(t)$ are converted into a soft importance distribution:
	\begin{equation}
		p(t)=\frac{\exp(\hat{s}(t)/\tau)}{\sum_j \exp(\hat{s}(j)/\tau)}
	\end{equation}
	where $\tau$ is a temperature parameter annealed during training.
	
	To avoid premature collapse of the importance distribution during early training, we include an entropy regularization term $H(p)$ that encourages exploration over frames when the selector is still uncertain.
	
	At inference time, LFS selects exactly $K$ frames using a stratified Top-$K$ strategy. The video timeline is divided into $K$ equal temporal segments, and one frame is selected from each segment:
	\begin{equation}
		t_i=\arg\max_{t\in \mathrm{segment}(i)} \hat{s}(t), \quad i=1,\ldots,K
	\end{equation}
	This strategy guarantees full temporal coverage and prevents temporal clustering. When applicable, the first and last frames are always retained.
	
	During training, gradients are propagated through the soft distribution $p(t)$, while the stratified Top-$K$ operator is applied only for hard frame selection at inference.
	
	\subsection{Caption-Guided Optimization}
	
	To align frame importance with detailed captioning quality, LFS is trained under the guidance of a frozen video captioner. Given temporal importance scores $p(t)$, we select a truncated set of top-ranked frames to control computation and normalize their weights to obtain $w$ with $\sum_t w_t = 1$. This truncated-and-renormalized design enables efficient optimization under a fixed frame budget while preserving differentiability.
	
	To encourage compact frame selection, we impose an $\ell_1$ regularization on $w$. Although $w$ is normalized, the $\ell_1$ penalty operates on the truncated candidate set prior to renormalization, discouraging mass spreading and promoting concentration on representative frames. In addition, an entropy regularizer is applied to $p(t)$ to encourage early exploration and is gradually annealed during training.
	
	We inject the frame weights $w$ into the captioner through a differentiable weighting hook at a designated visual module. Let $F_t \in \mathbb{R}^{L \times D}$ denote per-frame visual features at the hooked layer. The hook computes a fused representation:
	\begin{equation}
		F^{\mathrm{fused}} = \sum_{t=1}^{T} w_t F_t ,
	\end{equation}
	which is broadcast back to match the original tensor shape. This operation acts as a global modulation signal, while the captioner’s internal temporal and linguistic modeling remains unchanged.
	
	The captioner input consists of a textual prompt concatenated with the ground-truth caption. We compute token-level cross-entropy only on caption tokens, masking prompt and padding tokens:
	\begin{equation}
		\mathcal{L}_{\mathrm{cap}}(w) =
		- \sum_{i \in \mathcal{C}}
		\log p_{\theta}\!\left(y_i \mid y_{<i}, F^{\mathrm{fused}}\right)
	\end{equation}
	where $\mathcal{C}$ indexes caption tokens. The captioner parameters $\theta$ are frozen, so gradients propagate only to the temporal scoring network through $w$.
	
	To reduce captioner bias and stabilize optimization, we adopt a relative caption objective by subtracting a uniform baseline computed over the same truncated frame set:
	\begin{equation}
		\mathcal{L}_{\mathrm{cap}}^{\mathrm{rel}}(w)
		=
		\mathcal{L}_{\mathrm{cap}}(w)
		-
		\mathcal{L}_{\mathrm{cap}}(w^{\mathrm{uni}})
	\end{equation}
	where $w^{\mathrm{uni}}$ assigns equal weights to the selected frames.
	
	The final training objective is:
	\begin{equation}
		\mathcal{L}
		=
		\mathcal{L}_{\mathrm{cap}}^{\mathrm{rel}}(w)
		+
		\lambda_0 \| w \|_1
		-
		\lambda_{\mathrm{ent}} H(p)
	\end{equation}
	where the $\ell_1$ term promotes compact frame weighting, the entropy regularizer encourages exploration, and the relative caption loss aligns frame selection with captioning quality.

	\subsection{Training Details}
	
	LFS is trained for five epochs using AdamW optimizer with a learning rate of $1\times10^{-4}$ and weight decay of $1\times10^{-4}$. We use mixed-precision training with a batch size of 1. The temperature parameter is annealed from $\tau_{\text{start}}=2.0$ to $\tau_{\text{end}}=1.0$. Sparsity and entropy regularization weights are set to $\lambda_0=0.01$ and $\lambda_{\mathrm{ent}}=0.01$, respectively, and the number of selected frames is capped at 16 per video. The frozen captioning model used for caption-guided supervision is Qwen3-VL-8B. In the training process of LFS, we employed three NVIDIA A800 GPUs.
	
	The training set consists of 1{,}588 videos of 2--3 minutes collected from WebVid-10 \cite{Bain21}, TGIF \cite{7780871}, Charades \cite{10.1007/978-3-319-46448-0_31}, YouCook2 \cite{8578472}, and TREC-VTT \cite {awad2023overviewevaluatedvideoretrieval}. Training captions are obtained via distillation from Qwen3-VL-253B-A22B. 
	
	\section{ICH-CC Benchmark}
	
	Existing video understanding benchmarks, such as VDC and Dream-1K, primarily prioritize evaluation challenges and thus leads to misalignment with human cognitive understanding ability. Therefore, we present a ICH-CC benchmark to reflect human cognitive ability. 
	
	As shown in Fig.4, five experienced annotators each carefully labeled 20 videos only, producing the detailed captions and 100 QA counterparts, and conducted five rounds of cross-checking to ensure that every item was reviewed through all annotators. All videos are sourced from real-world business scenarios in the context of intangible cultural heritage Chinese cuisine. 
	\begin{figure}[htbp]
		\centering
		\includegraphics[width=0.5\textwidth]{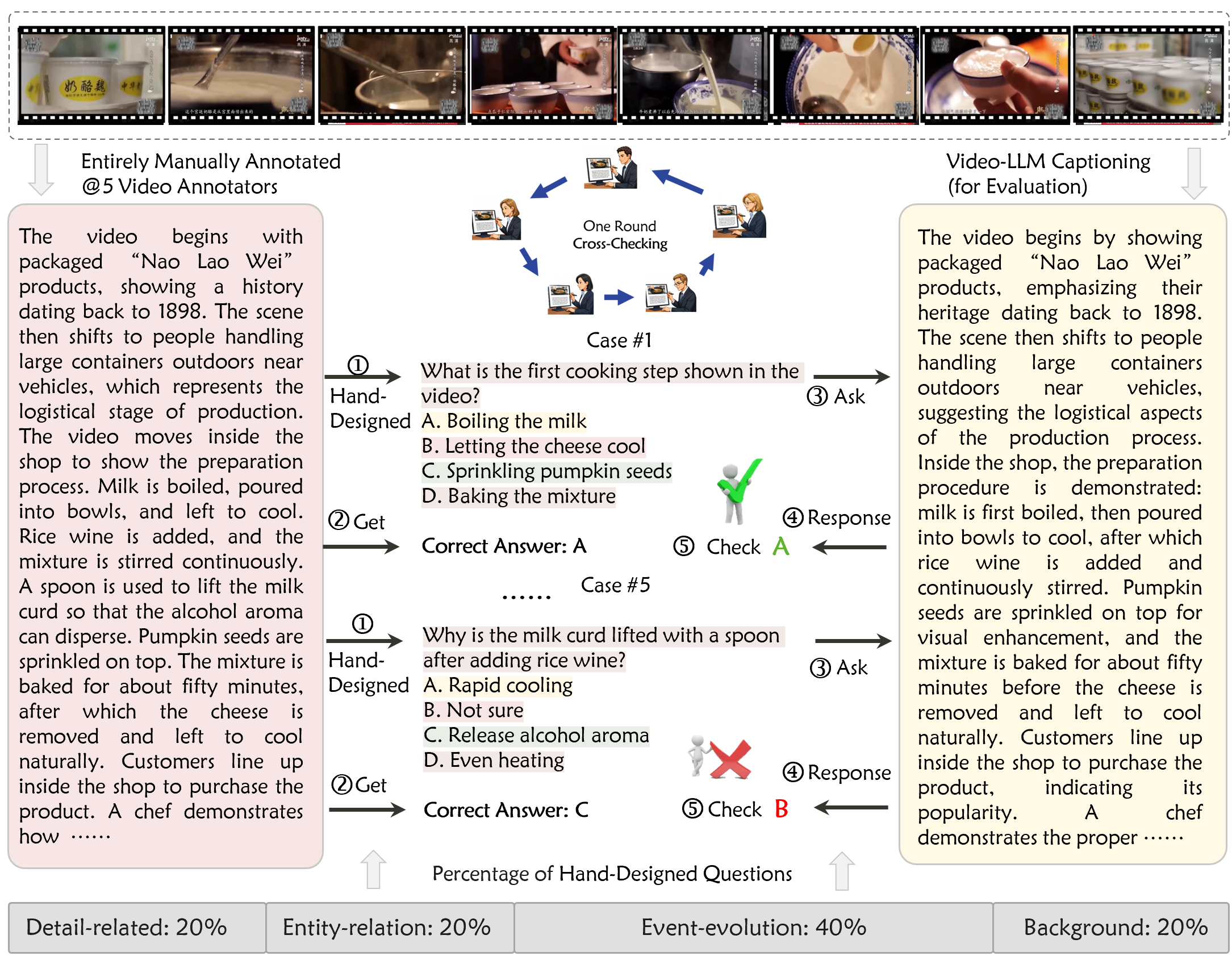}
		\caption{Overview of ICH-CC construction pipeline. The five annotators cross-checked each other’s annotated data. After five rounds, each data was thoroughly examined by all annotators to eliminate human biases.}
		\label{fig:ichcc_pipeline}
	\end{figure}
	
	ICH-CC consists of two subsets, ICH-CC-en for English and ICH-CC-zh for Chinese, with each containing 500 evaluation questions designed from 100 videos (2-3 minutes each). ICH-CC has a balanced composition of question types: detail-related, entity-relation, event-evolution, and background questions account for 20\%, 20\%, 40\%, and 20\%, respectively. 
	
	Models are tasked with generating a detailed caption for each video, which is then assessed by answering associated objective questions derived from human-authored descriptions. A question is deemed correct if the caption provides sufficient evidence to support the correct answer. The final score is calculated based on the percentage of correctly answered questions, with a focus on both semantic coverage and temporal consistency.

	\begin{figure*}[t]
		\centering
		\includegraphics[width=\textwidth]{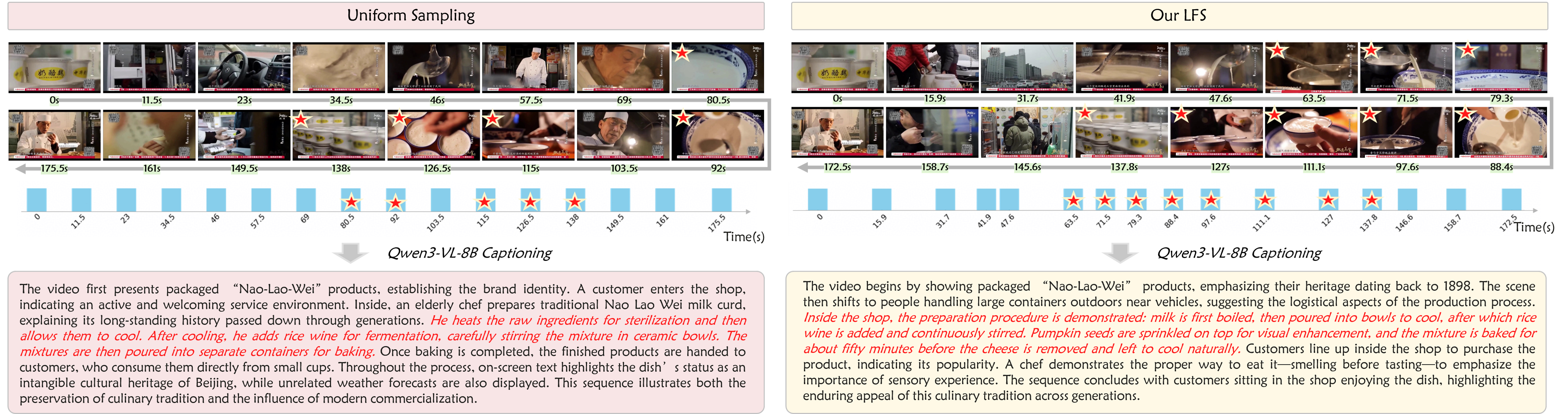}
		\caption{Qualitative results on the ICH-CC-en benchmark. The left shows Qwen3-VL-8B using uniform sampling and our LFS. \ding{73} indicates selected event-aware frames, and the bar chart visualizes their distribution along the timeline.}
		\label{fig:sample}
	\end{figure*}

	\section{Experimental Results} 
	
	\subsection{Analysis of ICH-CC benchmark}
	
	\textit{Quantitative Analysis.} Table~1 reports results on the ICH-CC benchmark, where ICH-CC-en/ICH-CC-zh denote accuracy on the English and Chinese subsets. Strong baselines such as Qwen2.5-VL-7B and Qwen3-VL-8B show competitive performance, with Qwen3-VL-8B reaching 74.25\% on ICH-CC-zh, outperforming prior open-source models.
	
	Integrating LFS yields consistent gains under a fixed budget of 16 frames. LFS + Qwen2.5-VL-7B improves overall accuracy from 68.46\% to 71.77\% (+3.31\%). LFS + Qwen3-VL-8B further achieves 75.05\% (+3.82\%), with gains of +4.47\% on ICH-CC-en and +3.14\% on ICH-CC-zh. These results demonstrate that our LFS with event awareness and temporal diversity effectively enhances detailed video captioning.
	
	\begin{table}[htbp]
		\begin{flushleft}
			\resizebox{0.48\textwidth}{!}{
				\begin{tabular}{l c c c}
					\hline
					\textbf{Model} & \textbf{ICH-CC-en} & \textbf{ICH-CC-zh} & \textbf{Overall} \\
					\hline
					Vicuna-v1.5-7B & 66.25 & 69.36 & 67.81 \\
					Video-ChatGPT-7B & 67.88 & 68.25 & 68.07 \\
					LLaMA-VID & 64.25 & 63.77 & 64.01 \\
					LongVA-7B & 66.37 & 65.89 & 66.13 \\
					ShareGPT4Video-8B & 62.36 & 65.96 & 64.16 \\
					MovieChat-7B & 62.38 & 62.86 & 62.62 \\
					Qwen2.5-VL-7B* & 67.52 & 69.39 & 68.46 \\
					Qwen3-VL-8B* & 68.20 & 74.25 & 71.23 \\
					\hline
					\textit{Our LFS} + Qwen2.5-VL-7B* & \textbf{69.88} & \textbf{73.65} & \textbf{71.77} \\
					$\Delta$Acc / $\Delta$Score & +2.36 & +4.26 & +3.31 \\
					\textit{Our LFS} + Qwen3-VL-8B* & \textbf{72.67} & \textbf{77.43} & \textbf{75.05} \\
					$\Delta$Acc / $\Delta$Score & +4.47 & +3.14 & +3.82 \\
					\hline
				\end{tabular}
			}
			\caption{Accuracy (\%) on the ICH-CC benchmark. ICH-CC-en and ICH-CC-zh represent the English and Chinese subsets, respectively. Bold numbers indicate the best scores. * indicates the baseline before applying LFS}
		\end{flushleft}
	\end{table}

	\begin{table*}[!t]
		\centering
		\resizebox{\textwidth}{!}{
			\begin{tabular}{p{4cm} >{\centering\arraybackslash}p{3cm} >{\centering\arraybackslash}p{3cm} >{\centering\arraybackslash}p{3cm} >{\centering\arraybackslash}p{3cm} >{\centering\arraybackslash}p{3cm}}
				\toprule[1.2pt]
				\textbf{Model} & \parbox{3cm}{\centering \vspace{10pt} \textbf{Camera} \\ \textbf{Acc / Sim} \vspace{10pt}} & \parbox{3cm}{\centering \vspace{10pt} \textbf{Short} \\ \textbf{Acc / Sim} \vspace{10pt}} & \parbox{3cm}{\centering \vspace{10pt} \textbf{Background} \\ \textbf{Acc / Sim} \vspace{10pt}} & \parbox{3cm}{\centering \vspace{10pt} \textbf{Main Object} \\ \textbf{Acc / Sim} \vspace{10pt}} & \parbox{3cm}{\centering \vspace{10pt} \textbf{Detailed} \\ \textbf{Acc / Sim} \vspace{10pt}} \\
				\midrule
				Gemini-1.5 Pro & 38.68 / 2.05 & 35.71 / 1.85 & 43.84 / 2.23 & 47.32 / 2.41 & 43.11 / 2.22 \\
				\midrule
				Vicuna-v1.5-7B & 21.68 / 1.12 & 23.06 / 1.17 & 22.02 / 1.15 & 22.64 / 1.16 & 23.09 / 1.20 \\
				LLaMA-VID & 39.47 / 2.10 & 29.92 / 1.56 & 28.01 / 1.45 & 31.24 / 1.59 & 25.67 / 1.38 \\
				Video-ChatGPT-7B & 37.46 / 2.00 & 29.36 / 1.56 & 33.68 / 1.70 & 30.47 / 1.60 & 24.61 / 1.26 \\
				MovieChat-7B & 37.25 / 1.98 & 32.55 / 1.59 & 28.99 / 1.54 & 31.97 / 1.64 & 28.82 / 1.46 \\
				VILA-7B & 34.33 / 1.83 & 30.40 / 1.55 & 35.15 / 1.80 & 33.38 / 1.72 & 29.78 / 1.58 \\
				Video-LLaVA-7B & 37.48 / 1.97 & 30.67 / 1.63 & 32.50 / 1.70 & 36.01 / 1.85 & 27.36 / 1.43 \\
				LLaVA-1.5-7B & 38.38 / 2.04 & 28.61 / 1.51 & 34.86 / 1.79 & 34.62 / 1.76 & 33.43 / 1.73 \\
				LongVA-7B & 35.32 / 1.90 & 31.94 / 1.63 & 36.39 / 1.85 & 40.95 / 2.11 & 27.91 / 1.48 \\
				ShareGPT4Video-8B & 33.28 / 1.76 & 39.08 / 1.94 & 35.77 / 1.81 & 37.12 / 1.89 & 35.62 / 1.84 \\
				InternVL-2-8B & 39.08 / 2.11 & 33.02 / 1.74 & 37.47 / 1.89 & 44.16 / 2.22 & 34.89 / 1.82 \\
				AuroraCap-7B* & 43.50 / 2.27 & 32.07 / 1.68 & 35.92 / 1.84 & 39.02 / 1.97 & 41.30 / 2.15 \\
				Qwen3-VL-8B* & 47.66 / 2.69 & 38.28 / 1.35 & 39.98 / 2.16 & 42.36 / 2.56 & 55.56 / 2.59 \\
				\midrule
				\textit{Our LFS} + AuroraCap-7B* & 44.10 / 2.35 & 34.57 / 1.77 & 36.02 / 1.98 & 40.65 / 2.77 & 43.04 / 2.21 \\
				$\Delta$Acc / $\Delta$Sim & +0.60 / 0.08 & +2.50 / 0.09 & +0.10 / 0.14 & +1.63 / 0.80 & +1.74 / 0.06 \\
				\midrule
				\textit{Our LFS} + Qwen3-VL-8B* & \textbf{48.82 / 2.97} & \textbf{39.58 / 1.75} & \textbf{41.62 / 2.59} & \textbf{43.59 / 2.87} & \textbf{57.58 / 2.71} \\
				$\Delta$Acc / $\Delta$Sim & +1.16 / 0.32 & +1.30 / 0.40 & +1.64 / 0.43 & +1.23 / 0.31 & +2.02 / 0.12 \\
				\toprule[1.2pt]
			\end{tabular}
		}
		\caption{Results of VDC benchmark. * represents the baseline models, the bold numbers represent the best scores in the open-source models. Acc and Sim indicates the accuracy and the similarity between the outputs and answers, respectively.}
	\end{table*}
	
	\textit{Qualitative Analysis.} Figure~5 compares uniform sampling and LFS on an ICH-CC video (Nai-Lao-Wei) with 16 frames. Uniform sampling captures only five event-aware frames and misses key short steps, while LFS retrieves eight frames covering the main procedure. The bar plots show that uniform sampling wastes frames on redundant intervals, whereas the stratified Top-$K$ strategy distributes samples across the timeline and concentrates on high-importance segments, preserving both early context and late-stage frames. Consequently, captions from LFS-selected frames recover more detailed procedures (boiling, cooling, adding rice wine, stirring, garnishing, and baking) than uniform sampling.

	\subsection{Analysis of VDC Benchmark}
	
	We evaluate our approach on VDC, which measures accuracy and similarity scores. Following AuroraCap, we prompt Llama-3.1-8B to answer based on the predicted captions. As shown in Table~2, integrating LFS with AuroraCap-7B and Qwen3-VL-8B yields consistent improvements across categories. In particular, the largest gains appear in Detailed, where accuracy improves from 41.30\% to 43.04\% for AuroraCap-7B and from 55.56\% to 57.58\% for Qwen3-VL-8B. These results align with our method design: event-aware scoring plus stratified selection better captures critical moments for detailed descriptions, and the improvements generalize across architectures.

	\subsection{Analysis of Dream-1K Benchmark}
	
	Table~3 summarizes results on Dream-1K. Compared with vanilla Tarsier2-7B using uniform sampling, Our LFS + Tarsier2-7B achieves the same F1 (0.40), with a slight recall improvement (0.48 vs.\ 0.47) and a marginal precision decrease (0.34 vs.\ 0.35). This similarity mainly arises from dataset characteristics: Dream-1K consists of very short clips (typically $<10$s), where 8 frames already provide near-complete coverage, leaving limited headroom for selection. The modest recall gain indicates LFS can still recover additional cues, while the overall effect reflects a recall–precision trade-off. Overall, Dream-1K serves as a short-video robustness test, showing LFS can be integrated into Tarsier2-7B without degrading performance, while its advantages are more evident on longer videos (VDC and ICH-CC).
	
	\begin{table}[htbp]
		\begin{flushleft}
			\tiny
			\resizebox{0.46\textwidth}{!}{
				\begin{tabular}{l c c c}
					\cline{1-4}
					\textbf{Model} & \textbf{Precision} & \textbf{Recall} & \textbf{F1} \\ 
					\cline{1-4}
					GPT-4V & 0.30 & 0.41 & 0.34 \\
					GPT-4o & 0.36 & 0.43 & 0.39 \\
					Gemini1.5 Pro & 0.35 & 0.38 & 0.36 \\
					\cline{1-4} 
					Video-LLaVA & 0.16 & 0.28 & 0.20 \\
					MiniGPT-4V & 0.22 & 0.26 & 0.24 \\
					LLaVA-NeXT-Video & 0.21 & 0.36 & 0.26 \\
					VideoChat2 & 0.23 & 0.31 & 0.27 \\
					PLLaVA-34B & 0.22 & 0.38 & 0.28 \\
					Tarsier2-7B* &  \textbf{0.35} & 0.47 & \textbf{0.40} \\ 
					\cline{1-4}
					\textit{Our LFS} + Tarsier2-7B* & 0.34 & \textbf{0.48} & \textbf{0.40} \\ 
					$\Delta$Precision / Recall / F1 & -0.01 & +0.01 & +0.00 \\
					\cline{1-4}
				\end{tabular}
			}
			\caption{Precision, Recall and F1 of Dream-1K benchmark. The numbers in bold indicates the best performance among the open-source models.}
		\end{flushleft}
	\end{table}
	
	\subsection{Zero-Shot Video Question-Answering}
	Table~4 reports video QA results where models answer questions solely from video descriptions, without access to raw frames, directly testing whether detailed captions support downstream reasoning. LFS consistently improves Qwen3-VL-8B across all benchmarks. On MVBench, accuracy increases from 68.7\% to 70.8\%, indicating improved capture of complex actions. On VideoMME without subtitles, LFS yields a +1.2\% gain, suggesting stronger visual grounding from captions alone. We also observe improvements on MLYU-MCQ (79.0\%) and VideoMMMU (66.8\%), both of which require fine-grained event understanding. These results confirm that LFS produces more informative and event-aware descriptions, enabling reliable zero-shot QA task.
	
	\begin{table}[htbp] 
		\begin{flushleft}
			\normalsize 
			\resizebox{0.48\textwidth}{!}{
				\begin{tabular}{l c c c c}
					\hline
					\parbox[c][28pt][c]{2.2cm}{\textbf{Model}} 
					& \parbox[c][28pt][c]{1.2cm}{\centering\textbf{MVBench}} 
					& \parbox[c][28pt][c]{2.1cm}{\centering\textbf{VideoMME}\\\centering w/o sub} 
					& \parbox[c][28pt][c]{1.3cm}{\centering\textbf{MLYU}\\\centering{MCQ}}
					& \parbox[c][28pt][c]{1.4cm}{\centering\textbf{Video}\\\centering{MMMU}} \\
					\hline
					GPT5-Nano & - & 49.4 & 52.6 & 40.2 \\
					Gemini2.5-Flash-Lite & - & 65.0 & 69.3 & 63.0 \\
					\hline
					Qwen2.5-VL-72B & 70.4 & \textbf{73.3} & 74.6 & 60.2 \\
					Qwen3-VL-4B & 68.9 & 69.3 & 75.3 & 56.2 \\
					Qwen3-VL-8B* & 68.7 & 71.4 & 78.1 & 65.3 \\
					\hline
					\textit{Our LFS} + Qwen3-VL-8B* & \textbf{70.8} & 72.6 & \textbf{79.0} & \textbf{66.8} \\
					$\Delta$Acc & +2.1 & +1.2 & +0.9 & +1.5 \\
					\hline
				\end{tabular}
			}
			\caption{Accuracy (\%) of video QA benchmark. The numbers in bold indicates the best performance.}
		\end{flushleft}
	\end{table}
	
	\subsection{Ablation Study}
	
	All ablations are conducted using Qwen3-VL-8B as the fixed captioning backbone to isolate the contribution of individual LFS components. Table~\ref{tab:ablation} reports results with a fixed frame budget of $K{=}16$. The full model achieves the best performance across all benchmarks (72.67\% on ICH-CC-en, 77.43\% on ICH-CC-zh, and 57.58\% on VDC Detailed), while removing any component degrades performance.
	
	\begin{table}[h]
		\centering
		\setlength{\tabcolsep}{5pt}
		\resizebox{0.48\textwidth}{!}{
			\begin{tabular}{lccc}
				\toprule
				\parbox[l][24pt][c]{3.2cm}{\textbf{Method (K=16)}} &
				\parbox[c][24pt][c]{2.0cm}{\centering\textbf{ICH-CC-en}\\\%} &
				\parbox[c][24pt][c]{2.0cm}{\centering\textbf{ICH-CC-zh}\\\%} &
				\parbox[c][24pt][c]{2.2cm}{\centering\textbf{VDC Detailed}\\\%} \\
				\midrule
				Uniform Sampling & 68.20 & 74.25 & 55.56 \\
				\midrule
				LFS w/o Stratified & 67.12 & 71.23 & 56.20 \\
				LFS w/o $H_2$ & 72.58 & 77.21 & 57.30 \\
				LFS w/o $\mathcal{L}_{\mathrm{cap}}$ & 71.53 & 77.10 & 57.20 \\
				\midrule
				LFS w/o gating & 72.41 & 77.23 & 56.98 \\
				LFS w/o norm & 72.34 & 76.36 & 56.42 \\
				\midrule
				\textbf{Full LFS} & \textbf{72.67} & \textbf{77.43} & \textbf{57.58} \\
				\bottomrule
			\end{tabular}
		}
		\caption{Ablation study of our LFS. 
			\textit{LFS w/o Stratified} removes the stratified mechanism, 
			\textit{LFS w/o $H_2$} removes event-level temporal modeling, 
			\textit{LFS w/o $\mathcal{L}_{\mathrm{cap}}$} removes caption-guided supervision, 
			\textit{LFS w/o gating} removes global gating, and 
			\textit{LFS w/o norm} removes normalization.}
		\label{tab:ablation}
	\end{table}
	Removing the stratified mechanism results in the largest drop (e.g., ICH-CC-zh: 77.43 $\rightarrow$ 71.23; VDC: 57.58 $\rightarrow$ 56.20), highlighting the importance of enforcing temporal coverage and avoiding clustering for long, stage-wise events. Removing event-level modeling ($H_2$) or caption-guided supervision ($\mathcal{L}_{\mathrm{cap}}$) also reduces performance, confirming the need to model coherent events and optimize selection for caption quality. Removing global gating or normalization causes smaller but consistent drops, indicating their role in stabilizing temporal scoring. Overall, these results validate that LFS benefits from the joint design of stratified selection, event-aware temporal modeling, and caption-guided supervision for accurate and temporally diverse detailed captioning.

	\section{Conclusion}
	We developed a learnable frame selector named LFS before video-LLMs by integrating event-aware temporal modeling, stratified Top-K selection, and caption-guided supervision for detailed video captioning. Experiments show that LFS consistently improves video-LLM backbones and produces more informative video descriptions, demonstrating the value of effective frame selection for scalable video understanding.
	
	\bibliographystyle{named}
	\bibliography{ijcai26}
	
\end{document}